\documentclass[review]{elsarticle}

\usepackage{lineno,hyperref}

\usepackage{times}
\usepackage{epsfig}
\usepackage{graphicx}
\usepackage{amsmath}
\usepackage{amssymb}

\usepackage{slashbox}
\usepackage{tabularx}

\usepackage{graphicx}
\usepackage{amsmath,amssymb} 
\usepackage{color}

\usepackage{amssymb}

\usepackage{url}

\usepackage{algorithm}               
\usepackage{algorithmic}            
\usepackage{multirow}                
\usepackage{xcolor}

\journal{Optics and Laser Technology}

\begin{document}

\begin{frontmatter}

\title{Single Reference Image based Scene Relighting \\ via Material Guided Filtering}


\author[BestiCS]{Xin Jin}
\author[BestiCS]{Yannan Li}
\author[UIBE]{Ningning Liu}
\author[BestiCS]{Xiaodong Li
\corref{mycorrespondingauthor}}
\ead{lxd@besti.edu.cn}

\author[BestiCS]{\\Xianggang Jiang}
\author[BestiEE]{Chaoen Xiao}

\author[CAS]{Shiming Ge
\corref{mycorrespondingauthor}}
\cortext[mycorrespondingauthor]{Corresponding authors}
\ead{geshiming@iie.ac.cn}

\address[BestiCS]{Department of Computer Science
and Technology,}
\address[BestiEE]{Department of Electronic Information Engineering,\\ Beijing Electronic Science and Technology Institute, Beijing, 100070, P.R. China}
\address[UIBE]{School of Information Technology
and Management,\\ University of International Business and Economics, Beijing, 100029, P.R. China}
\address[CAS]{Institute of Information Engineering,\\ Chinese Academy of Sciences, Beijing, 100093, P.R. China}

%
%
%
%

\begin{abstract}
Image relighting is to change the illumination of an image to a target illumination effect without known the original scene geometry, material information and illumination condition. We propose a novel outdoor scene relighting method, which needs only a single reference image and is based on material constrained layer decomposition. Firstly, the material map is extracted from the input image. Then, the reference image is warped to the input image through patch match based image warping. Lastly, the input image is relit using material constrained layer decomposition. The experimental results reveal that our method can produce similar illumination effect as that of the reference image on the input image using only a single reference image.
\end{abstract}


\end{frontmatter}


\section{Introduction}

Image relighting is a hot topic in the communities of computer vision, image processing and computational photography. The applications of image relighting include visual communication, film production and digital entertainment, etc. Image relighting is to change the illumination of an image to a target illumination effect without known the original scene geometry, material information and illumination condition. Comparing with face, object and indoor scene, two main challenges are for outdoor scene relighting: (1) large-scale outdoor scene with multiple objects, which are not easy to reconstruct; (2) complex illumination in outdoor scene, which is hard to be controlled manually. Recently, reference image based image relighting has shown great potential \cite{ShihSiggraph2013} \cite{HaberCVPR2009} \cite{ChenTIP2013} \cite{PeersSiggraph2007}. Currently, for face relighting, the reference images are changed from multiple and a pair \cite{ChenECCV2010} to a single \cite{ChenTIP2013}. For object relighting \cite{HaberCVPR2009} \cite{JinICVRV2016} and scene relighting \cite{ShihSiggraph2013}, multiple or a pair reference images are still needed \cite{LuCEE2017} \cite{LuAccess2017} \cite{LuOptSoc2015} \cite{LuIEICE2016} \cite{ZhouPR2016}.

We propose a novel outdoor scene relighting method, which needs only a single reference image and is based on material constrained layer decomposition. Firstly, the material map is extracted from the input image. Then, the reference image is warped to the input image through patch match based image warping. Lastly, the input image is relit using material constrained layer decomposition. The experimental results reveal that our method can produce similar illumination effect as that of the reference image on the input image using only a single reference image.


\section{Scene Relighting}
\subsection{Method Overview}
\begin{figure}
\centering
\includegraphics[width=\textwidth]{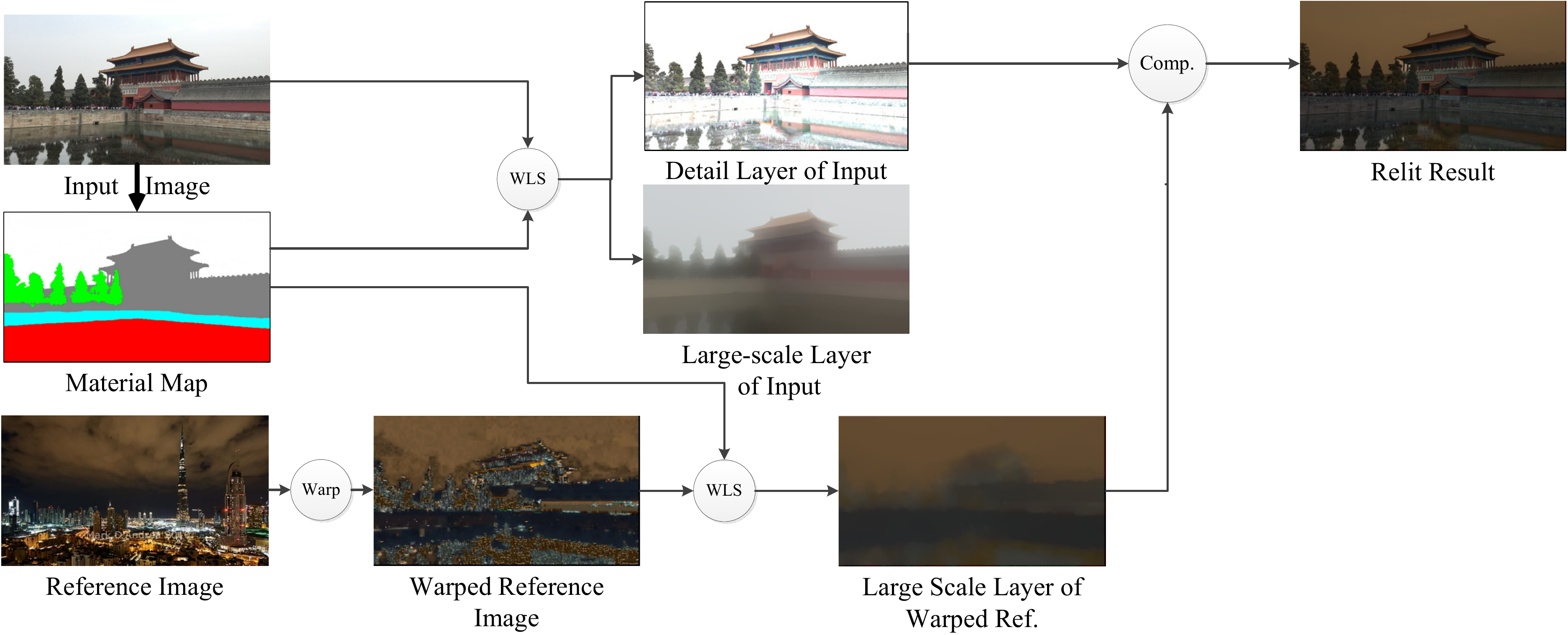}
\caption{Scene relighting using only a single reference image.}
\label{fig:overview}
\end{figure}

Our proposed method can be divided into 4 steps, as shown in Fig. \ref{fig:overview} (1) the input image is segmented to the material map using the method of Bell et al. \cite{BellCVPR2015}. Every pixel of the material map is assigned by a material label; (2) the reference image is warped to the structure of the input image by the patch match warping; (3) each channel of the input image and the reference is decomposed to large-scale layer and detail layer under material constrain; (4) the final relit results are obtained by composing the details of the input image and the large-scale layer of the warped reference image.

\subsection{Material Segmentation}
\begin{figure}
\centering
\includegraphics[width=12cm]{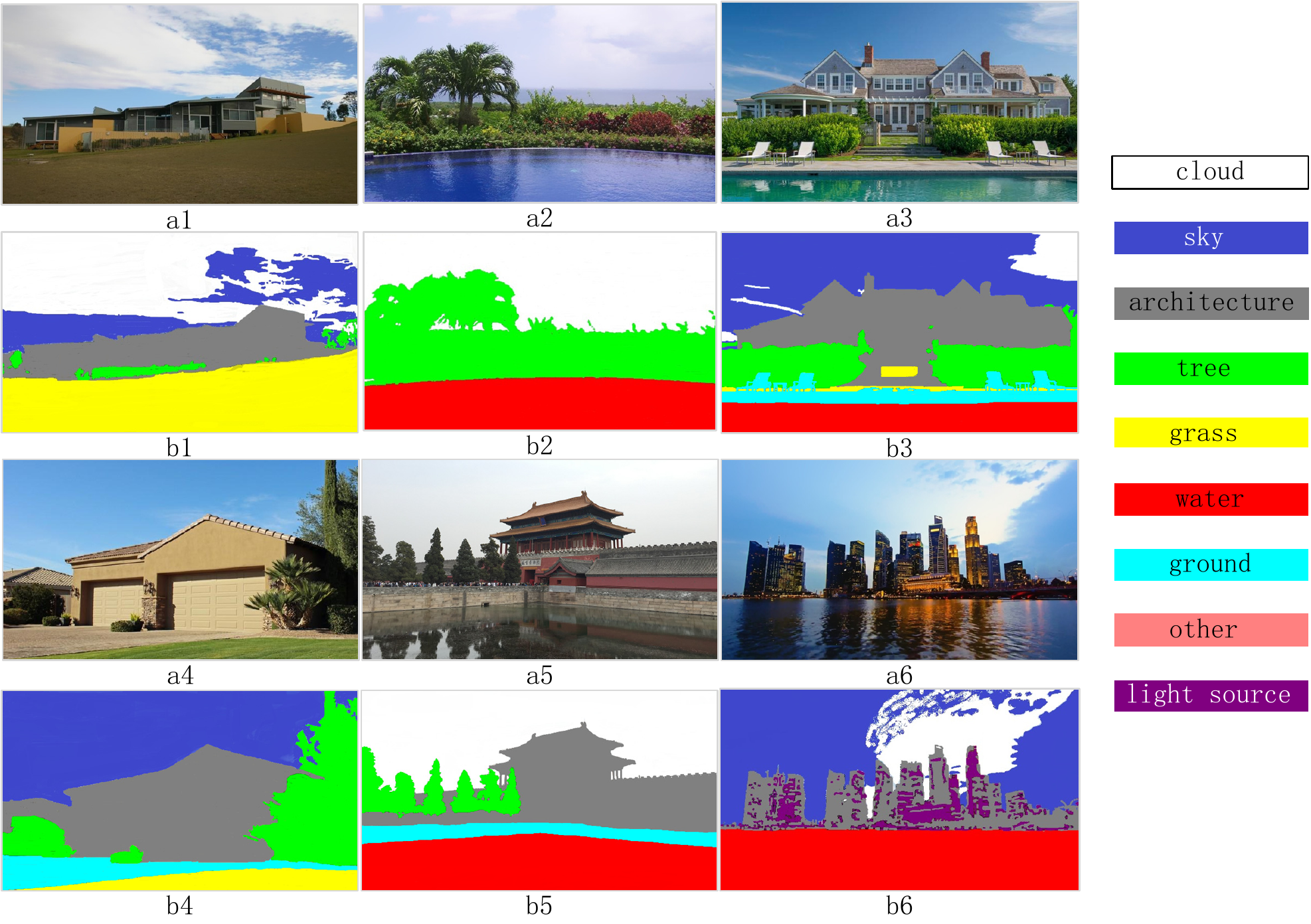}
\caption{The material maps of the input images.}
\label{fig:material}
\end{figure}

The input image is segmented according to the material of each pixel. We use the method of Bell et al. \cite{BellCVPR2015} to obtain material label of each pixel. We make the material segmentation because that in different material region, different relighting operations should be conducted. We select 9 sorts of materials, which often appear in outdoor scene images, as shown in Fig. \ref{fig:material}. We recolor each pixel according to the material label to get the material map. The first and the third lines are the input images. The second and the forth lines are the corresponding material maps.

\subsection{Reference Image Warping}
In face image relighting, the reference face image can be warped by face landmark detection / face alignment. However, in outdoor scene, we cannot find such similar structure easily. The outdoor scene contains multiple objects. Thus, we use the patch match method to warp the reference image to the input image, i.e. to align the reference and the input image. The patch match algorithm is similar as the method of Barnes et al. \cite{BarnesECCV2010}. We use the neighbor patches whose best matched patches have already been found to improve matching result of current patch. The difference from Barnes et al. \cite{BarnesECCV2010} is that we use 4 neighbor patches instead of 3 ones. 

The basic idea is to find the most similar patch in the reference image to substitute the original patch in the input image. Two basic assumptions are made: (1) the matched patches of the neighbor patches in the input image are mostly neighbor; (2) large scale random search region may also contain matched patch.

We denote the input image as $A$ and the reference image as $B$. The coordinate of a patch is represented as coordinate of the left up corner of the patch. The Nearest Neighbor Field (NNF) is defined as $\operatorname{f}$ , whose definition domain is the coordinates of all the patches in $A$. The value of the NNF is the offset of the coordinate of matched patch in $B$. We denote the coordinate of the original patch in $A$ as $a$ and the coordinate of the matched patch in $B$ as $b$, then:

\begin{equation}
\operatorname{f}(a)= b-a
\end{equation}
The distance between the original patch and the matched patch is defined as $\operatorname{D}(v)$, which describes the distance between the patch $a$ in $A$ and patch $a+v$ in $B$. The distance is computed by the Euclidean distance \cite{LoweIJCV2004}. The warping method contains three steps: initialization, propagation and random search.

\begin{figure}
\centering
\includegraphics[width=11cm]{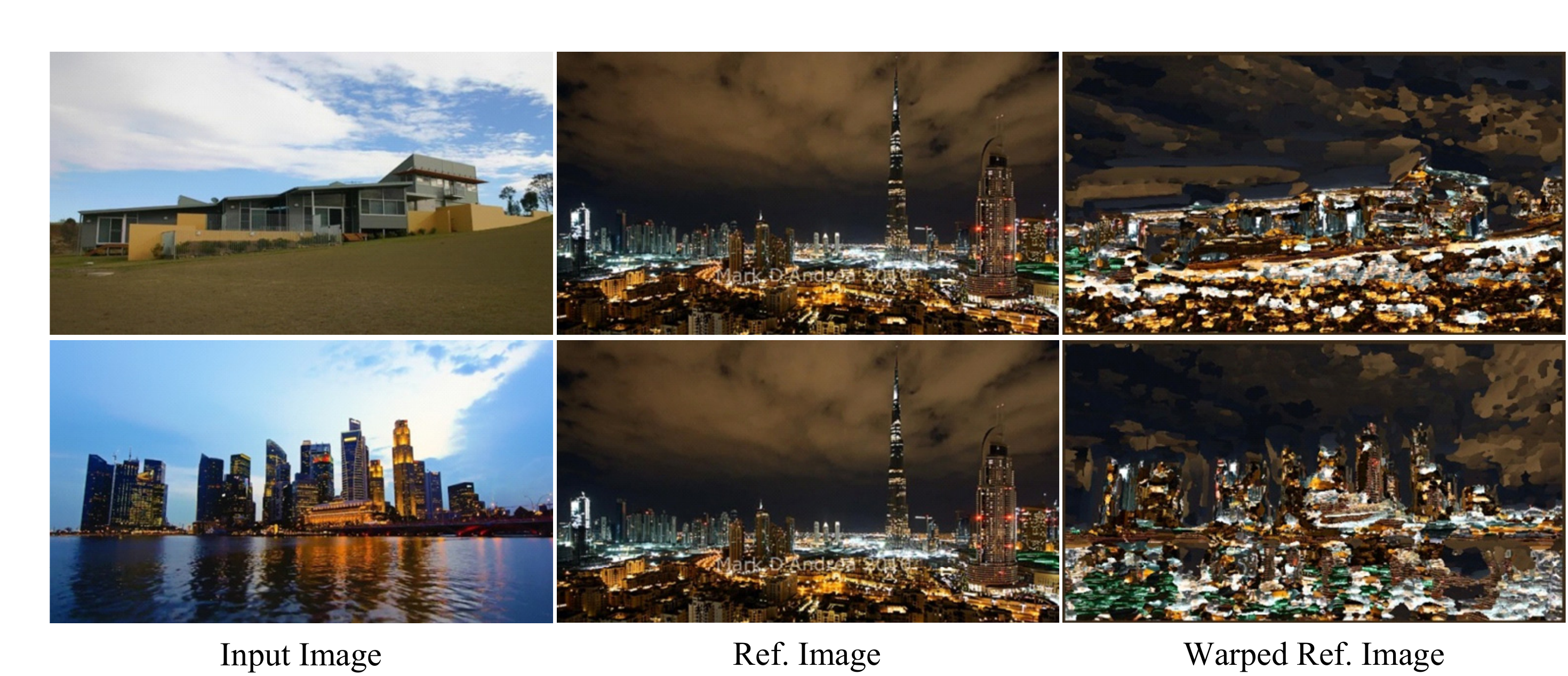}
\caption{Patch match based scene image warping.}
\label{fig:warp}
\end{figure} 

\begin{itemize}

\item \textbf{Initialization}. The initial offset of each patch in $A$ is randomized around the patch.

\item \textbf{Propagation}. As assumed above, the matched patches of the neighbor patches in the input image are mostly neighbor. We use the neighbor patches whose best matched patches have already been found to improve matching result of current patch. The $\operatorname{f}(x-1,y)$, $\operatorname{f}(x,y-1)$ and $\operatorname{f}(x-1,y-1)$ are used:

\begin{equation}
\operatorname{f}(x,y) = \operatorname{min} \{\operatorname{D}(\operatorname{f}(x,y)),\operatorname{D}(f(x-1,y)),\operatorname{D}(\operatorname{f}(x,y-1)),\operatorname{D}(\operatorname{f}(x-1,y-1))\}
\end{equation}

\item \textbf{Random Search}. As assumed above, large scale random search region may also contain matched patch. We use a search window whose size is declined exponentially.

\begin{equation}
u_i=v_0+w\alpha^i R_i,
\end{equation}
where $v_0=f(x,y)$, $R_i$ is a random point in $[1,1]×[-1,1]$. $w$ is the max search radius. $\alpha$ is the declining rate of the radius.

\end{itemize}
The warped results of some reference images are shown in Fig. \ref{fig:warp}.

\subsection{Layer Decomposition and Composition}
We use the WLS filter \cite{FarbmanSiggraph2008} to decompose image into large-scale layer and detail layer, which can be considered as the illumination component and non-illumination component. Using the large-scale layer of the warped reference to substitute the large-scale layer of the input can produce the final relit result. The outdoor scene contains various objects with various materials. Thus for different material, different decomposition parameters should be used.  Each channel $l$ of the input image and the reference image is filtered to a large-scale layer s. The detail layer $d$ is obtained by:

\begin{eqnarray}
d = l/s.
\label{eq:detail}
\end{eqnarray} 

The original WLS filter uses the same smoothness level over the whole image. When using the WLS filter for our scene relighting task, we need make regions with different materials with different smooth levels. Thus, we set different smoothness levels in regions with different materials. We modified the original WLS \cite{FarbmanSiggraph2008} as:



\begin{equation}
\label{eq:wls1}
\centering
E = |l-s|^2+\lambda H(\nabla s,\nabla l)
\end{equation}

\begin{equation}
\label{eq:wls2}
H(\nabla s,\nabla l) = \sum_p ( \lambda(p) ( \frac{(\partial s / \partial x)^2_{p}}{(\partial l / \partial x)^{\alpha}_{p} + \epsilon} + \frac{(\partial s / \partial y)^2_{p}}{(\partial l / \partial y)^{\alpha}_{p} + \epsilon} ) ),
\end{equation}
where, ${|l-s|}^2$ is the data term, which is to let $l$ and $s$ as similar as possible, i.e., to minimize the distance between $l$ and $s$. $H(\nabla s,\nabla l)$ is the regularization (smoothness) term, which makes $s$ as smooth as possible, i.e. to minimize the partial derivative of $s$. $p$ is the pixel of the image. $\alpha$ controls over the affinities by non-linearly scaling the gradients. Increasing $\alpha$ will result in sharper preserved edges. $\lambda$ is the balance factor between the data term and the smoothness term. Increasing $\lambda$ will produce smoother images. $\epsilon$ is a very small number, so as to avoid the division by $0$. Our $\lambda$ is the smoothness level constrained by different materials, using the material map derived in Section 3.2:

\begin{equation}
\lambda=\nabla l+ \operatorname{gray}(l_m )/255,
\end{equation}
where, $\nabla l$ is the gradient of $l$. $l_m$ is the material map of $l$, and the gray is the gray value of $l_m$:

\begin{equation}
\operatorname{gray}=(R \times 0.2989 + G \times 0.587 + B \times 0.114)
\end{equation}
The minimization of Eq. (1) and Eq. (2) can be solved by the off-the-shell methods such as Lischinski [17]. At last, using the large-scale layer of the warped reference to substitute the large-scale layer of the input can produce the final relit result.

\section{Experimental Results}
In this section, we show the experimental results of our proposed method and the comparison with the state of the art method.

\subsection{The Scene Relit Results}
\begin{figure}
\centering
\includegraphics[width=\textwidth]{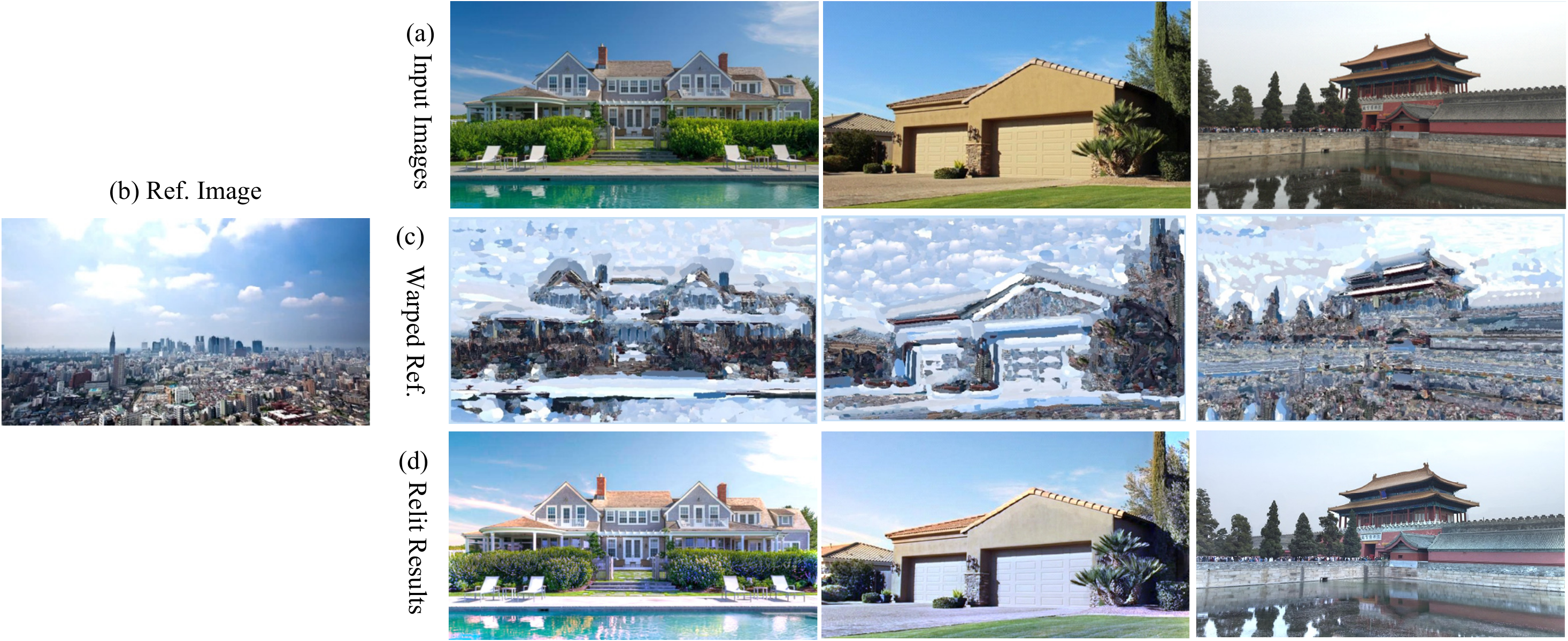}
\caption{The relit results of our method.}
\label{fig:relit}
\end{figure}

The relit results of our method are shown in Fig. \ref{fig:relit}. (a): multiple input images, (b): the same reference image, (c): warped reference image, (d): relit results of input images using (c), The experimental results reveal that, the relit input image have similar illumination effect as that of the reference image.

\subsection{Comparison with Other Methods}
\begin{figure}
\centering
\includegraphics[width=\textwidth]{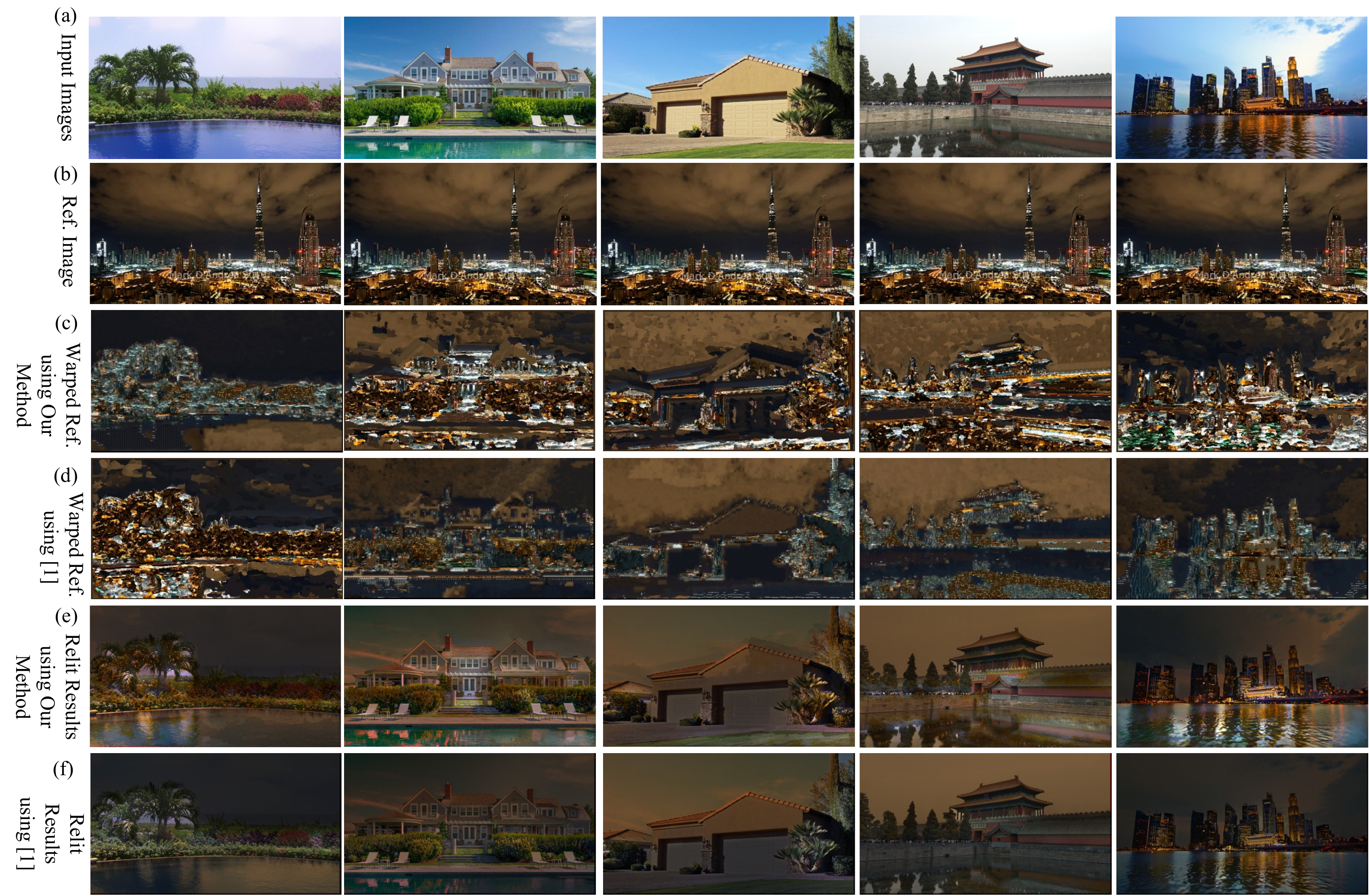}
\caption{Comparison with \cite{ShihSiggraph2013}.}
\label{fig:comparison}
\end{figure}

We compare our method with the state of the art method \cite{ShihSiggraph2013}, which needs a time-lapse video captured by a fixed camera working for 24 hours. While our method needs only a single reference image. We randomly select 5 input images for comparison. As shown in Fig. \ref{fig:comparison}, (a): multiple input images, (b) the reference image, (c): warped reference images to corresponding input images, (d): warped reference images using method of [1], note that they need a time-lapse video for warping, (e): the relit results using our proposed method, (f): the relit results using the method of [1].The results reveal that our method can produce similar relit results as those of [1], with only a single reference image.

\section{Conclusions}
We propose a novel outdoor scene relighting method, which needs only a single reference image and is based on material constrained layer decomposition. The experimental results reveal that our method can produce similar illumination effect as that of the reference image on the input image using only a single reference image.

\section*{Acknowledgements}
We thank all the reviewers and PCs. This work is partially supported by the National Natural Science Foundation of China (Grant NO.61402021, 61401228, 61640216), the Science and Technology Project of the State Archives Administrator (Grant NO. 2015-B-10), the open funding project of State Key Laboratory of Virtual Reality Technology and Systems, Beihang University (Grant NO. BUAA-VR-16KF-09), the Fundamental Research Funds for the Central Universities (Grant NO.2016LG03, 2016LG04), the China Postdoctoral Science Foundation (Grant NO.2015M581841), and the Postdoctoral Science Foundation of Jiangsu Province (Grant NO.1501019A).


\bibliography{mybibfile}

\end{document}